\documentclass[10pt,twocolumn,letterpaper]{article}

\usepackage{cvpr} % uncomment this for the final submission
\usepackage{times}
\usepackage{epsfig}
\usepackage{graphicx}
\usepackage{amsmath}
\usepackage{amssymb}
\usepackage{booktabs}
\usepackage{lipsum}
\usepackage{hyperref}
\usepackage{listings}
\usepackage{float}
\usepackage{array}
\usepackage{tikz}

\floatstyle{plaintop}
\newfloat{lstfloat}{tb!}{lop}
\floatname{lstfloat}{Algorithm}

\lstdefinestyle{algorithm} 
    {
        basicstyle = \small\ttfamily,
        breaklines = true,
}

\def\dag{\textsuperscript{\textdagger}}
\def\ast{\textsuperscript{*}}
\def\ddag{\textsuperscript{\textdaggerdbl}}
\def\sec{\textsuperscript{\S}}

\newcommand\copyrighttext{%
  \footnotesize \textcopyright 2025 IEEE. Personal use of this material is permitted. Permission from IEEE must be obtained for all other uses, in any current or future
  media, including reprinting/republishing this material for advertising or promotional purposes, creating new collective works, for resale or redistribution to servers or lists, or reuse of any copyrighted component of this work in other works.}
\newcommand\copyrightnotice{%
\begin{tikzpicture}[remember picture,overlay]
\node[anchor=south,yshift=10pt] at (current page.south) {\fbox{\parbox{\dimexpr\textwidth-\fboxsep-\fboxrule\relax}{\copyrighttext}}};
\end{tikzpicture}%
}

\begin{document}

%%%%%%%%% TITLE
\title{First International StepUP Competition for Biometric Footstep Recognition: Methods, Results and Remaining Challenges}

\author{Robyn Larracy\ast, Eve MacDonald\ast, Angkoon Phinyomark\ast, Saeid Rezaei\dag, Mahdi Laghaei\ddag,\\
Ali Hajighasem\sec, Aaron Tabor\ast, and Erik Scheme\ast \\[1.0ex]
\ast University of New Brunswick, Canada\\
{\tt\small \{rlarracy,eve.macdonald,aphinyom,aaron.tabor,escheme\}@unb.ca}\\[1.0ex]
\dag University College Cork, Ireland\\
{\tt\small saeid.rezaei@ucc.ie}\\[1.0ex]
\ddag Islamic Azad University, Iran\\
{\tt\small m.laghaei@iau.ac.ir}\\[1.0ex]
\sec University of New South Whales, Australia\\
{\tt\small a.hajighasem@student.unsw.edu.au}\\
}

\maketitle
\copyrightnotice
\thispagestyle{empty}

%%%%%%%%% ABSTRACT
\begin{abstract}

Biometric footstep recognition, based on a person's unique pressure patterns under their feet during walking, is an emerging field with growing applications in security and safety. 
However, progress in this area has been limited by the lack of large, diverse datasets necessary to address critical challenges such as generalization to new users and robustness to shifts in factors like footwear or walking speed. 
The recent release of the UNB StepUP-P150 dataset, the largest and most comprehensive collection of high-resolution footstep pressure recordings to date, opens new opportunities for addressing these challenges through deep learning. 
To mark this milestone, the First International StepUP Competition for Biometric Footstep Recognition was launched. 
Competitors were tasked with developing robust recognition models using the StepUP-P150 dataset that were then evaluated on a separate, dedicated test set designed to assess verification performance under challenging variations, given limited and relatively homogeneous reference data.
The competition attracted global participation, with 23 registered teams from academia and industry. 
The top-performing team, Saeid\_UCC, achieved the best equal error rate (EER) of 10.77\% using a generative reward machine (GRM) optimization strategy. 
Overall, the competition showcased strong solutions, but persistent challenges in generalizing to unfamiliar footwear highlight a critical area for future work.

%Design new techniques for classifying footsteps under real-world variations, pushing the boundaries of biometric footstep recognition and shaping the future of this emerging field.

\end{abstract}

%%%%%%%%% BODY TEXT
\section{Introduction}

Footstep recognition is a behavioural biometric modality that identifies or verifies individuals based on the unique pressure patterns under their feet during walking. 
Pressure-sensitive flooring is used to capture complex gait characteristics directly during natural movement, offering inherent robustness to environmental factors such as lighting and occlusion, and requiring no active cooperation or deliberate presentation from the user.
This emerging biometric modality presents a promising and convenient solution for access control and monitoring for transportation hubs, critical infrastructure, and other high-security environments. 
Early work in this area demonstrates promise, with impressive recognition accuracies in controlled laboratory environments \cite{Pataky2012, Horst2023,Derlatka2023,Xie2020}. 
%Each footstep recording is a dynamic and information-dense representation of gait, capturing spatial and temporal characteristics that are unique to each individual. 

However, key considerations for real-world implementation remain underexplored, particularly regarding robustness to natural variability in gait. 
Footstep recordings can exhibit high within-subject variation due to factors such as cognitive load, walking speed, footwear, or carried items \cite{Oh2017,Connor2018}. 
Since enrollment data is typically collected under brief, controlled conditions (e.g., one shoe type, one walking speed), these variations can introduce substantial domain shifts during deployment. 
Consequently, recognition systems must generalize to unseen and unpredictable conditions while maintaining reliability against unenrolled individuals, despite having only a small and homogeneous set of reference samples.
This is an open and ongoing challenge that demands novel solutions. 
While recent advances in deep learning have made it increasingly feasible to address these issues (demonstrated by recent advancements in robust in-the-wild vision-based gait recognition \cite{Shen2024}), progress has been slowed by the lack of a large, diverse public dataset to support development and benchmarking. 
Existing datasets containing underfoot pressure or force data during walking, such as CASIA-D \cite{zheng2011evaluation}, GaitRec \cite{horsak2020gaitrec}, and SFootBD \cite{vera2012comparative} have not been sufficient for this use case because of factors including a limited number of participants and sources of variability, the use of single-point measure force plates, and/or a lack of comprehensive data labeling.

The recent release of UNB’s StepUP-P150 dataset \cite{Larracy2025} marks a major milestone for the field, offering a resource of approximately 200,000 high-resolution footsteps from 150 individuals under varied footwear and walking speed conditions. 
To celebrate this release and accelerate progress toward real-world deployment, the First International StepUP Competition for Biometric Footstep Recognition was launched. 
Competitors were tasked with developing robust recognition models capable of tackling two key challenges on a dedicated and withheld test set of footsteps: (1) generalizing to new users with limited reference data, and (2) adapting to novel conditions, including unseen footwear and walking speeds. 
As the first competition in pressure-based footstep biometrics, the StepUP challenge aimed to establish a foundational benchmark and provide a platform for advancing the state-of-the-art. 
The competition attracted global participation and a diverse array of methodological approaches, with the winning solution, submitted by team Saeid\_UCC, achieving an equal error rate (EER) of 10.77\% on the challenging evaluation set. 
Nonetheless, consistent trends across submissions highlight ongoing challenges, particularly in generalizing to unfamiliar personal footwear, underscoring the need for continued innovation.

\section{Datasets}

\begin{figure}[b!]
    \centering
    \includegraphics[width=\linewidth]{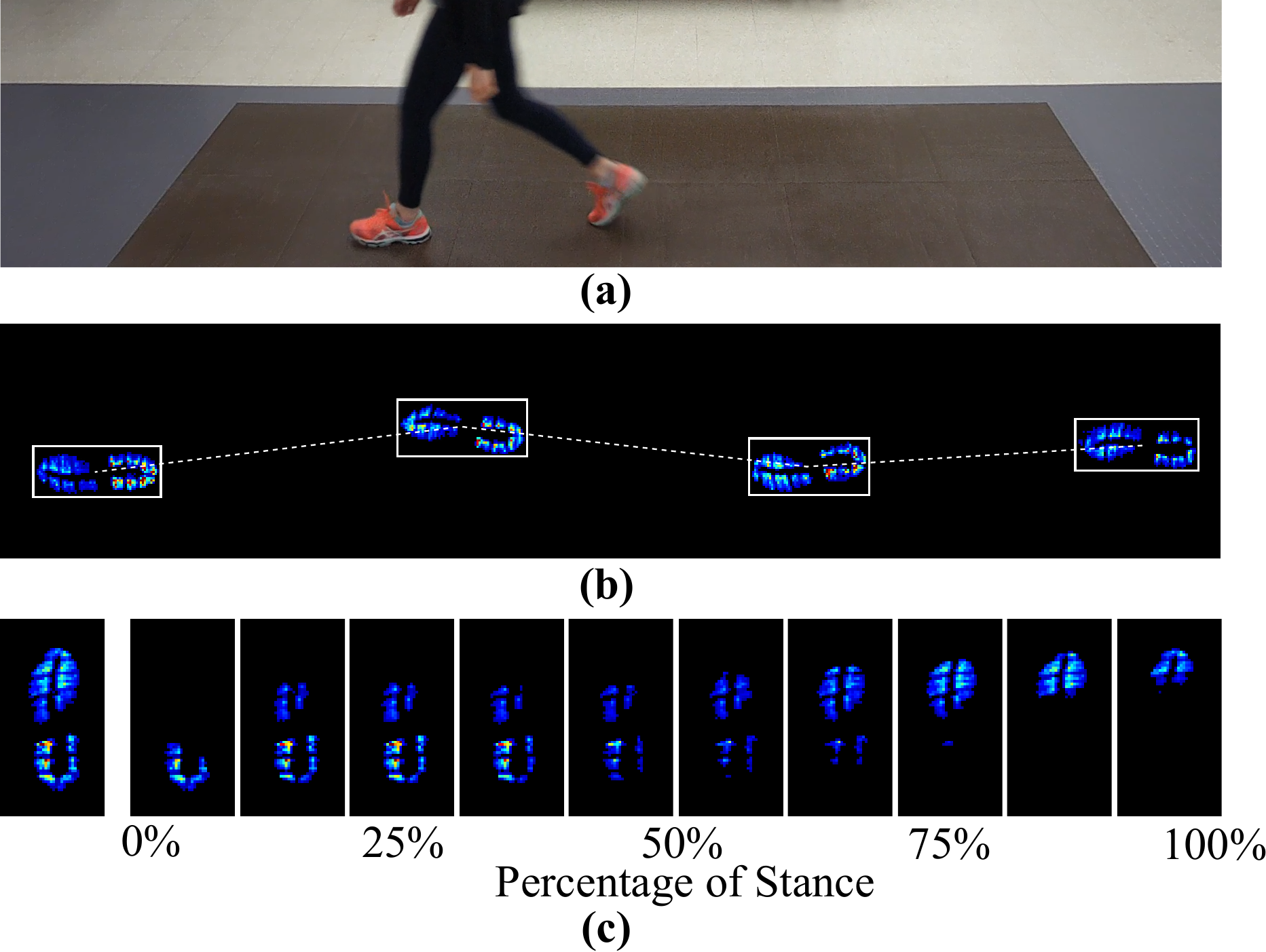}
    \caption{Overview of the collection procedure and example footstep recordings: (a) a participant walking across the pressure sensor-instrumented platform in a pair of their own personal shoes, (b) a series of footstep pressures captured during a pass across the platform, and (c) an example of an extracted and pre-processed footstep recording, shown at several time points throughout the stance.}
    \label{fig:example_step}
\end{figure}

Competitors were provided three distinct datasets for the competition (Table 1), which contained footstep recordings collected during the same 18-month study at the University of New Brunswick (REB 2022-132).
During the study, participants walked back and forth across a commercial 3.6 m \texttimes~1.2 m platform instrumented with a high-resolution grid of piezoresistive sensors (4 sensors/cm$^2$, 100 Hz; Stepscan Technologies Inc.).
Each participant completed sixteen 90-second walking trials, consisting of four different footwear conditions (BF: barefoot or sock-foot, ST: standard sneakers provided by the research team, and P1/P2: two pairs of personal shoes) and four different walking speeds (W1: preferred speed, W2: slowing to a stop, W3: slow, W4: fast).
The recordings were processed to extract individual footsteps, which were aligned to a common coordinate space and size. 
A combination of automated and manual methods was used to annotate each footstep (e.g., as left/right, incomplete recordings, and outliers).
This resulted in a set of labeled 3D pressure maps for each trial, with each footstep represented by a 101 $\times$ 75 $\times$ 40 (time, height, width) tensor.
Full details of the collection protocol and data processing steps are provided in \cite{Larracy2025}, and Fig.~\ref{fig:example_step} shows an example of a participant walking across the instrumented tiles along with their corresponding pressure recordings. 

\begin{table}[t!]
\caption{Overview of the three datasets used for the competition. The reference and probe sets were comprised of samples from individuals not included in the training set.}
\setlength{\tabcolsep}{9pt}
\begin{tabular}{@{}lccc@{}}
\toprule
\textbf{Dataset}   & \textbf{Users} & \textbf{Samples} & \textbf{Conditions}                           \\ \midrule
Training            & 150 & 200,000 & 4 speeds, 4 footwear \\
Reference & 15     & 150  & 1 speed, 1 footwear \\
Probe & 30           & 10,000             & 3 speeds, 3 footwear \\ \bottomrule
\end{tabular}
\end{table}

\subsection{Training Samples}
The training data consisted of the full set of recorded footsteps (all sixteen footwear \texttimes~speed combinations) from 150 individuals, which are fully annotated and publicly available for download as the UNB StepUP-P150 dataset \cite{Larracy2025}. 
The participants included 74 males and 76 females with an average age of 34 years (19-91 years), with race/ethnicities self-identified as White ($N = 106$), Middle Eastern ($N = 15$), South Asian ($N = 10$), East/Southeast Asian ($N = 11$), Black ($N = 1$), multi-ethnic ($N = 5$) or unknown/unspecified ($N = 2$). 
The dataset also encompasses a broad range of personal shoe types, with a total of 300 unique pairs including athletic shoes, dress shoes, boots, and sandals. 
In total, the StepUP-P150 dataset contains more than 200,000 pre-processed footsteps (approx. 1400 per participant) along with detailed metadata for each participant (e.g., age, sex, height, weight, foot size, descriptions of personal footwear) and each sample (e.g., left/right, foot rotation angle, location on tile grid, walking direction).  
These were made available to competitors for training models capable of robust feature extraction and downstream classification on the separate reference and probe sets.
% pipeline 1 preprocessing

\subsection{Reference and Probe Samples}
The reference and probe samples were drawn from a separate pool of thirty participants reserved specifically for the competition.
Data from these participants were excluded from the StepUP-P150 dataset following collection due to experimental deviations, missing data, hardware malfunction, or non-consent for video collection (as videos were used to verify annotations and data integrity in the public database). 
Only samples verified as complete and valid, however, were included for consideration in this competition. 
The thirty participants included 17 female and 13 male individuals with ages ranging 19-73 years and an average of 32.6 years.
The participants identified as White ($N = 19$), Middle Eastern ($N = 3$), South Asian ($N = 3$), East/Southeast Asian ($N = 1$), Aboriginal ($N = 1$), multi-ethnic ($N = 2$), or with unknown or unspecified race/ethnicity ($N = 1$).

%% reference samples
Fifteen of these participants were randomly selected as enrolled users, and labeled reference footsteps were provided as enrollment data.
For each user, these reference footsteps included five left footsteps and five right footsteps from the participant wearing a pair of their own personal shoes and walking at their preferred pace.   
This is indicative of the set of samples that could be collected during a realistic enrollment session lasting approximately 10-20 seconds.  
Minimal metadata was provided; each footstep was only labeled with the user's ID and whether the footstep was captured from the left or right foot. 
In total, the reference set consisted of 150 footsteps (10 per user).

%% Probe samples
The probe set comprised 10,000 footsteps used to evaluate verification performance, each labeled with a claimed identity corresponding to one of the users enrolled in the reference set. 
Of these, 30\% were true claims and 70\% were false claims. 
True-claim probes included footsteps recorded in both known conditions (i.e., footwear and walking speeds represented in the reference set) and unknown conditions. 
These unknown conditions included the users' second pair of personal shoes, the shared standard sneakers, and slow and fast walking speeds; these samples made up approximately 90\% of the true-claim probes and represented varying levels of domain shifts from the reference samples. 
False-claim probes consisted of footsteps from the other enrolled users (\textit{known impostors}) and the remaining 15 held-out users (\textit{unknown impostors}), again with a variety of footwear and speed conditions. 
Probes were randomly sampled and assigned, resulting in a diverse set drawn from 30 individuals and spanning nine walking condition combinations (three speeds and three footwear types; note that the barefoot and slow-to-stop trials were not included in the reference or probe sets).
Similar to the reference set, only the claimed identities and left and right labels were provided as metadata for the probe footsteps. 
Ground truth labels for the probe set were withheld to ensure fair evaluation. 
% not randomized uniformly: more likely for each shoe type to be assigned to one of the enrolled users. (less obvious for clustering).  
% approx equal proportions of three shoe types in both matched and non-matched
% more fast walking samples than slow or preferred

\section{Competition}
\subsection{Competition Task}
Competitors were tasked with performing verification for each of the 10,000 probe samples; that is, determining whether a given probe footstep matched its claimed identity. 
To encourage participation, contestants were only required to submit their model predictions, rather than a full source code solution. 
Each submission required two text files: (1) a file containing 10,000 similarity scores in a range between 0 and 1, each one corresponding to a probe sample, (2) a file containing a single numeric value representing the score threshold for match/non-match decisions. 
The competitors were not informed of the composition of the probe set (e.g., the proportion of true and false claims, or the footwear and speed conditions included), only that it included a mix of seen and unseen conditions and individuals.  
Competitors were free to use any method for developing their solutions based on the training and reference samples, with no restrictions on computational resources.
Although external data were permitted, to the best of the organizers’ knowledge, none of the top competitors opted to use additional datasets.

\subsection{Management and Starter Kit}
The competition was hosted on CodaBench \cite{codabench}, where competitors could upload their results, receive immediate performance feedback, and appear on a live leaderboard\footnote{CodaBench site: \url{www.codabench.org/competitions/5872}}.
Team sizes up to six members were permitted; however, many competitors chose to compete individually. 
Submissions were accepted between April and June of 2025, and teams could submit a maximum of five times per day up to a total of 100 submissions. 
%Competitors could freely use the labeled training and reference samples to develop their solutions, but were prohibited from using the probe samples for model tuning.

% starter kit implementation
Along with the three pre-processed datasets, competitors received a Python-based code starter kit upon registration. 
This implementation provided utilities for data loading, sample visualization, and exporting results, as well as a baseline solution for the competition task.
This example solution used a residual (2+1)D convolutional neural network, R(2+1)D CNN \cite{Tran2018}, which factorizes 3D convolutions into separate spatial (2D) and temporal (1D) components. 
The model was trained with a supervised contrastive loss to learn a discriminative embedding space from the sixteen footwear and walking speed combinations in the training set. 
Following feature extraction with this trained network, similarity scores were computed for the probe samples using cosine distance and cohort-based score normalization.
The full implementation is publicly available on GitHub\footnote{GitHub: \url{github.com/UNB-StepUP/stepup-starter-kit}}. 
This relatively na\"ive baseline was intended as a foundation for experimentation, offering competitors a simple starting point for developing improved solutions.

\subsection{Evaluation Metrics}

Solutions were primarily ranked based on equal error rate (EER); the decision threshold at which the rates of false matches and false non-matches were equal.
This metric was chosen for its ability to provide a balanced assessment of system performance with respect to the two types of errors.
However, to offer further insight and resolve ties if needed, several supplementary performance metrics were also calculated. 
These included the false match rate (FMR), false non-match rate (FNMR), overall accuracy (ACC), and balanced accuracy (BACC; computed as $100\% - (\rm{FMR} + \rm{FNMR})/2$), which were evaluated at the decision threshold submitted by the team along with their similarity scores.
The FMR100 was also computed, representing the FNMR when the decision threshold is fixed to achieve a FMR equal to 1\%.

\subsection{Participation}
The competition was promoted through the conference website, machine learning forums, social media platforms (e.g., LinkedIn), and targeted emails to researchers in gait biometrics and related fields. 
It attracted international interest from both academia and industry, with twenty-three teams registering from across North America, Europe, Asia, and Australia. 
Of these, twelve teams submitted solutions, with more than 400 submissions in total. %(median XX).
% many submissions representing refinements to hyperparameters, etc.

\section{Submitted Solutions} % Approaches
The competition organizers invited the winning teams to provide a description of their approaches. 
Descriptions provided by the top three teams are presented here.

%%%%%%%%%%% FIRST PLACE ALGORITHM

\subsection*{Saeid\_UCC (1st Place)} 
\noindent \textbf{Team}: Saeid Rezaei, \textit{University College Cork, Ireland}

\lstset{numbers=left, numberstyle=\tiny, stepnumber=1, numbersep=5pt}
\begin{lstfloat*}[tb!]
\vskip -8pt
\caption{Overview of the GRM hyperparameter optimization technique used for the winning solution proposed by Saeid\_UCC.}
\begin{lstlisting}[
    style = algorithm,
    mathescape=true,
    xleftmargin=0.35 cm,
    ]
// Phase 1: Offline Reward Model Estimation
Let $C$ be the set of model architecture classes (states).
Initialize dataset $D_c \leftarrow \emptyset$ for each class $c \in C$.
for each architecture class $c \in C$ do
    for $k$ = 1 to NUM_SAMPLES do
        Sample a random hyperparameter configuration $a_k$.
        Train model $c$ with hyperparameters $a_k$ to completion.
        Obtain the training log $L_k$ and true validation performance $\{P_{true}, k\}$.
        Extract dynamics feature vector $v_k$ from $L_k$.
        $D_c \leftarrow D_c \cup {(v_k, P_{true}, k)}$.
    end for
    Train a regression model on $D_c$ to learn the weight vector $W_c$ such that $P_{true} \approx W_c \cdot v$.
end for

// Phase 2: Online RL-based Hyperparameter Search
Initialize Q-table $Q(s, a)$ for all states $s \in C$ and actions $a \in A$.
for episode = 1 to MAX_EPISODES do
    Select an architecture class $s \in C$.
    Choose action (hyperparameters) $a$ from $Q(s, \cdot)$ using an exploration strategy.
    Train model $s$ with hyperparameters $a$ for a limited number of epochs.
    Obtain the partial training log $L_{episode}$.
    Extract dynamics feature vector $v_{episode}$ from $L_{episode}$.
    // Use the pre-learned model-specific estimator to generate the reward
    $reward \leftarrow W_s \cdot v_{episode}$.
    Update $Q(s, a)$ using the calculated reward.
end for
return The optimal policy $\pi*(s) = \arg\max_a Q(s, a)$.

\end{lstlisting}
\label{alg:GRM}
\vskip -14pt
\end{lstfloat*}

The dual challenge of selecting an optimal neural network architecture and tuning its corresponding hyperparameters represents a significant bottleneck in the deep learning pipeline. 
Traditional methods using grid search are computationally prohibitive and fail to leverage information gathered during the optimization process. 
In this work, we implemented a novel framework based on a Generative Reward Machine (GRM) to create an autonomous agent that intelligently navigates the joint space of architectures and hyperparameters. 
The methodology was applied across a diverse set of modern architectures, including Vision Transformer (ViT), Swin Transformer, ConvNeXt, and R(2+1)D, showcasing its adaptability. 
The GRM agent learns to interpret early training dynamics in an architecture-specific context, enabling it to identify high-performing configurations and prune unpromising trials. 
This state-aware, adaptive approach significantly enhances the sample efficiency of the search process, providing a robust solution for automated model selection and optimization.
The GRM is motivated by problems where the partial observations are inherently noisy, and so a perfect reward is not possible. 
An estimator maps observations of the environment states to a space of estimated states. 
In the simple case, the estimated space, the estimator, and the reward machine are all specified by the programmer. 
In the most complex case, the estimated space, the estimator, and the reward machine must be learned via interactions with the environment—for example, inferring a complex NN encoder for the estimator, mapping to an inferred latent space of estimated states.

This work introduces a GRM-based framework to automate the simultaneous selection of a model architecture and its hyperparameters. 
By decomposing the problem into an offline estimation phase and an online search phase, the framework creates an agent that learns to allocate computational resources efficiently, accelerating the discovery of optimal model-configuration pairs.
The proposed framework integrates a reward-estimation phase with a reinforcement learning (RL)-based search policy. 
It comprises two primary stages as detailed in Algorithm \ref{alg:GRM}. 
Phase 1 involves offline learning of architecture-specific reward models. 
Phase 2 uses these models for an efficient online RL-based search.

\noindent\textbf{Phase 1: Offline Reward Model Estimation}

The first phase aims to build an ``estimator'' for each potential model architecture class. 
This estimator learns to predict the final, true performance of a model based on the characteristics of its initial training dynamics. 
This preempts the need to train every configuration to completion during the search phase.
The state space $s$ for the RL agent is defined by the set of candidate architecture classes $C$ = \{ViT, Swin, ConvNeXt, R(2+1)D\}. 
For each class $c \in C$, a dataset is constructed by training the architecture with a diverse, randomly sampled set of hyperparameter configurations. 
For each complete training run, two key pieces of information are stored: the ground-truth validation performance ($P_{true}$) and the full training log ($L$), which contains metrics like per-epoch average loss. 
From each log, a feature vector ($v$) is extracted to numerically represent the training dynamics (e.g., rate of loss decrease, variance of terminal batch losses).
A regression model is then trained on this dataset for each architecture class to learn a weight vector $W_c$. This vector creates a mapping $P_{true} \approx W_c \cdot v$, effectively creating an expert model that can predict the final performance of a given architecture by observing only its initial training behavior.

\noindent\textbf{Phase 2: Online RL-based Hyperparameter and Architecture Search}

In the second phase, a Q-learning agent is deployed to perform an efficient online search. 
The agent's task is to learn an optimal policy, $\pi*(s, a)$, that selects the best action $a$ (a hyperparameter vector) for a given state $s$ (an architecture class).
The search proceeds episodically. 
In each episode, the agent selects an architecture $s$ and a corresponding set of hyperparameters $a$ based on its current Q-table, balanced by an exploration strategy like $\epsilon$-greedy. 
The selected model $s$ is then trained for a limited number of epochs -- just enough to generate a partial training log.
From this partial log, a dynamics feature vector $v_{episode}$ is extracted. 
The crucial step is the reward calculation: the reward signal passed to the agent is $reward = W_s \cdot v_{episode}$, where $W_s$ is the specific weight vector learned for that architecture during Phase 1. 
This architecture-aware reward allows the agent to make nuanced judgments, correctly interpreting that the early signs of a promising run for a ViT might differ from those of a ConvNeXt. 
The agent's Q-table is then updated with this reward, refining its policy for subsequent episodes. 
This process directs the search towards architecture-hyperparameter pairs that demonstrate favorable early-stage dynamics, reducing computational cost.
%%%%%%%%%%%%%%%%%%%%%%%%%%%%%%%%%%%%%%%%%%%%%%%%%%%%%%%
\begin{table}[b!]
\vskip -4pt
\centering
\caption{The model and hyperparameters chosen by the GRM framework proposed by the 1st place team, Saeid\_UCC.}
\begin{tabular}{w{c}{0.45\linewidth}w{c}{0.45\linewidth}}
\toprule
\textbf{Parameter} & \textbf{Value} \\
\midrule
Model & R(2+1)D \\
Optimizer & Adam \\
Learning Rate & 0.001 \\
Batch Size & 128 \\
Layer Sizes & (3, 3, 3, 3, 3) \\
Loss Function & TripletMarginLoss \\
Sample Size & (32, 32, 32) \\
Normalizer & minmax \\
UMAP Distance & 0.8 \\
Horizontal Flip & False \\
Vertical Flip & True \\
Rotation & True \\
Gradient Clip Norm & 10.0 \\
\bottomrule
\end{tabular}
\label{tab:model_config}
\end{table}
%%%%%%%%%%%%%%%%%%%%%%%%%%%%%%%%%%%%%%%%%%%%%%%%%%%%%%
This novel GRM process resulted in the architecture and hyperparameters shown in Table \ref{tab:model_config}.

%%%%%%%%%%% SECOND PLACE ALGORITHM 

\subsection*{Peneter ML (2nd Place)} 
\noindent \textbf{Team}: Mahdi Laghaei, \textit{Islamic Azad University, Iran}

To mitigate the computational cost of hyperparameter optimization (HPO) for deep spatio-temporal networks, we propose a Transfer-Initialized Multi-Fidelity Bayesian Optimization (TI-MFBO) framework. 
See Algorithm S1 in the Supplementary Material for an overview of the framework.
This approach enhances the sample efficiency of conventional Bayesian optimization by integrating two key innovations: transfer learning for surrogate model initialization and a multi-fidelity evaluation scheme. 
The core challenge in Bayesian optimization is the initial ``cold start'' phase, where the surrogate model lacks data on the objective function's landscape. 
To circumvent this, our method pre-trains the surrogate model on data from cheaper proxy tasks, such as optimizing a simpler 2D-CNN on max-intensity projections of the gait data or using heavily down-sampled input resolutions (sample\_sz). 
This transfer-learning step provides a highly informative prior, effectively ``warm-starting'' the optimization process for the full-fidelity model.

The operational mechanism of TI-MFBO relies on a sophisticated surrogate model, typically a Gaussian Process (GP), whose prior distribution is conditioned on the posterior learned from the proxy task. 
This ensures that the initial exploration of the target hyperparameter space is guided by relevant experience. 
The optimization proceeds using a multi-fidelity structure where computational resources are dynamically allocated. 
Low-fidelity evaluations—defined by training on a subset of participants or for a reduced number of epochs (n\_epochs) -- are used to rapidly discard unpromising hyperparameter configurations. 
This is governed by a cost-aware acquisition function, such as Multi-Fidelity Expected Improvement, which explicitly models the cost of evaluation at each fidelity level. 
A promotion logic, akin to successive halving, ensures that only configurations demonstrating sufficient promise at lower, cheaper fidelities are promoted for evaluation on higher, more expensive ones.

By synergizing a knowledgeable starting point with resource-aware evaluation, TI-MFBO substantially reduces the requisite number of expensive training cycles. 
This enables a more thorough and efficient search of the hyperparameter space, making it exceptionally well-suited for complex and resource-intensive deep learning models. 

The final result of the TI-MFBO approach used the provided R(2+1)D network with filter counts of (4,4,4,4,4,4). For optimization, the Adam optimizer was employed with an initial learning rate of 0.005.

%%%%%%%%%%% THIRD PLACE ALGORITHM 

\subsection*{CyberTI (3rd Place)} 
\noindent \textbf{Team}: Ali Hajighasem, \textit{University of New South Wales, Australia}

Traditional HPO treats the training process as a static black box. We introduce an alternative paradigm, Evolutionary Curriculum Co-Optimization (ECCO), which simultaneously optimizes both the model’s hyperparameters and the training data curriculum itself.
Algorithm S2 in the Supplementary Material provides an overview of the approach.
This method is predicated on the insight that the optimal hyperparameters may be dependent on the very strategy used to present data to the model. 
ECCO employs an evolutionary algorithm where each individual in the population represents a complete learning strategy. 
A genome in this population encodes two distinct sets of parameters: (1) the model’s hyperparameters, such as learning rate, optimizer, and layer sizes (lr, optimizer, layer\_sizes); and (2) the curriculum schedule. 
The curriculum is defined as a sequence of stages with increasing data complexity. 
For instance, an initial stage may use only data from simple walking conditions (e.g., barefoot (BF) at a single speed (W1)), with subsequent stages progressively introducing more challenging footwear (ST, P1, P2) and speed (W2, W3, W4) variables.

To operationalize this, the curriculum portion of the genome is represented as a list of tuples, [(epoch\_1, conditions\_1), (epoch\_2, conditions\_2), ...], dictating when to expand the dataset. The evolutionary process leverages specialized genetic operators; for instance, simulated binary crossover for continuous hyperparameters like learning rate, and uniform crossover for categorical choices like the optimizer. Mutations can involve slight perturbations of numerical values or shifts in the curriculum’s epoch boundaries. The fitness of each individual is evaluated based on a dual objective that balances final validation performance with computational cost, thereby rewarding strategies that are both effective and efficient. Through generations of selection, crossover, and mutation, ECCO evolves synergistic combinations of curricula and hyperparameters, capable of uncovering powerful and efficient training pathways that are inaccessible to methods that optimize these components in isolation.

Two deep learning models, a Convolutional Neural Network-Long Short-Term Memory (CNNLSTM) and a R(2+1)D, were used. %The R(2+1)D network architecture was identified as the superior model, achieving the highest performance score.
The optimal model configuration used the provided R(2+1)D Network, with filter counts of (3, 4, 4, 3, 4) across the network's spatiotemporal convolutional layers. 
The model used a TripletMargin loss function, designed to learn discriminative features by ensuring that an anchor sample is closer to a positive sample than it is to a negative sample by a specified margin. 
The AdamW optimizer was employed with an initial learning rate of 0.002. 
A learning rate scheduler was utilized to adjust the learning rate during training; while the specific scheduler was not explicitly defined by the algorithm's output, a conventional approach such as a step-wise decay or a reduce-on-plateau strategy is typically employed with such optimization schemes. 
The model was trained with a batch size of 128 for 100 epochs. 
Prior to being fed into the network, input samples were resized to a uniform dimension of (48, 48, 48), representing the temporal, height, and width dimensions, respectively.

%% Removed! Includes results
%The efficacy of this approach is demonstrated by the results of our evolutionary search, detailed in Figure 1. The top-performing configuration discovered by ECCO, identified by ID 278811, achieved a state-of-the-art Equal Error Rate (EER) of 11.5\%. This represents a significant 23\% reduction in error compared to the poorest-performing configuration (ID 276166, EER of 14.93\%) explored during the search, underscoring the method's ability to navigate the complex, combined search space and identify superior solutions.

\section{Results and Discussion}
%%%%%%%%%%%%%%%%%%%%%%%%%%%%%%%%%%%%%%%%%%%%%%%%%%%%%%%%%%%

\begin{table}[tb!]
\setlength{\tabcolsep}{1.5pt}
\caption{Performance metrics (in \%) for the top five ranked teams. While EER and FMR100 were computed using optimized decision thresholds, the remaining metrics (ACC, BACC, FNMR, FMR) were computed at the decision threshold submitted by the team.}
\begin{tabular}{@{}llcccccc@{}}
\toprule
\# & Team Name & EER & FMR100 & ACC & BACC & FNMR & FMR \\ \midrule
1 & Saeid\_UCC      & 10.77 & 59.63 & 88.51 & 88.94 & 10.0  & 12.13 \\
2 & Peneter ML   & 11.33 & 67.34 & 87.61 & 88.37 & 9.73  & 13.53 \\
3 & CyberTI      & 11.50 & 67.12 & 86.99 & 88.17 & 8.87  & 14.79 \\
4 & Anonymous  & 11.67 & 80.14 & 87.94 & 88.24 & 11.00 & 12.51 \\
5 & Technologist & 12.23 & 77.46 & 87.36 & 87.75 & 11.27 & 13.23 \\ 
\bottomrule
\end{tabular}
\label{tab:leaderboard}
\end{table}
%%%%%%%%%%%%%%%%%%%%%%%%%%%%%%%%%%%%%%%%%%%%%%%%%%%%%%%%%%%

\begin{figure}[b!]
    \centering
    \includegraphics[width=\linewidth]{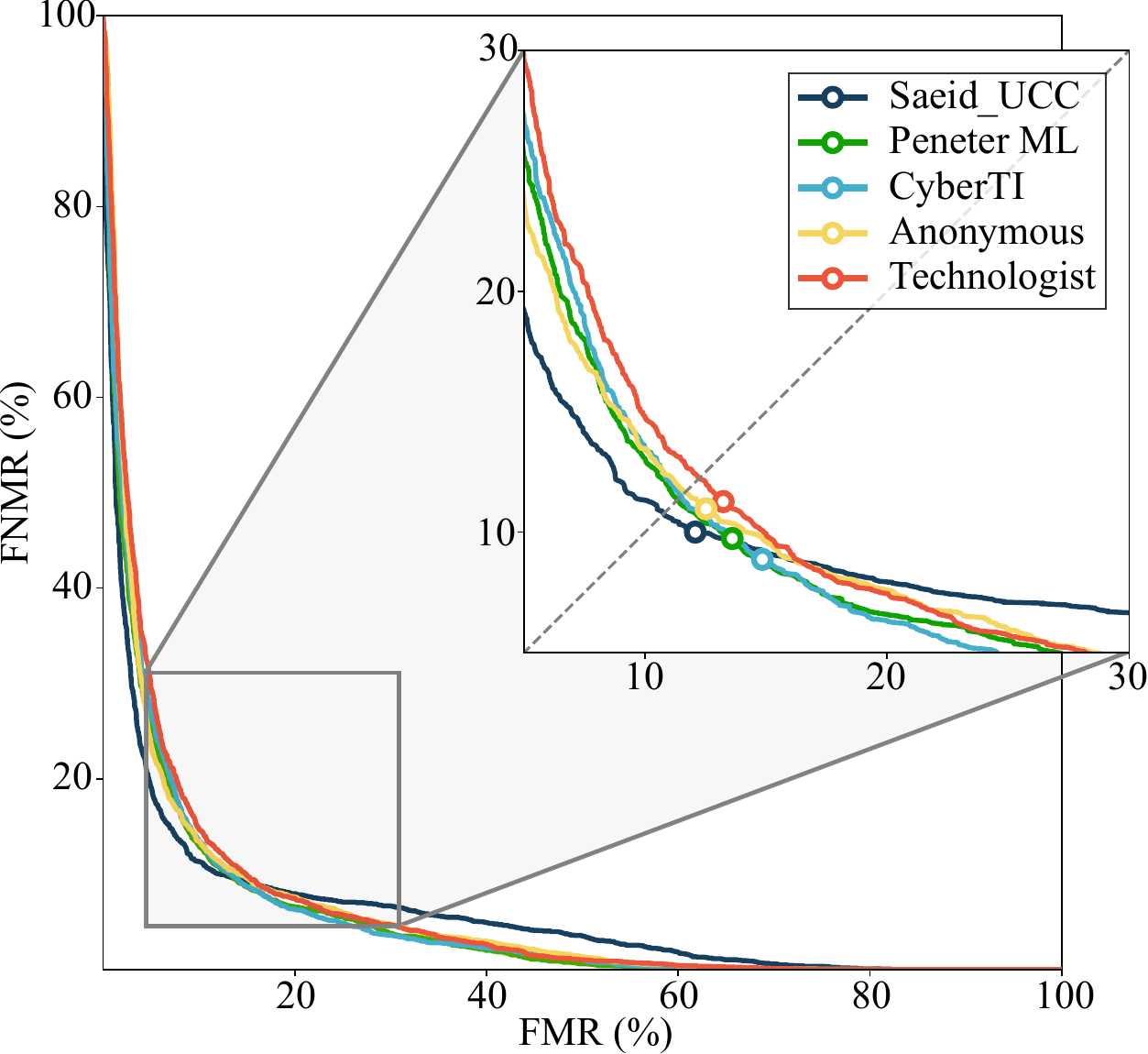}
    \caption{Detection error tradeoff (DET) curve for the top five solutions. Circle markers represent model performance at the competitor's submitted threshold, while EERs are denoted by each curve's intersection with the dotted line (FNMR = FMR).}
    \label{fig:DET}
\end{figure}

%%%%%%%%%%%%%%%%%%%%%%%%%%%%%%%%%%%%%%%%%%%%%%%%%%%%%%%%%%%
\begin{figure}[tb!]
    \centering
    \includegraphics[width=\linewidth]{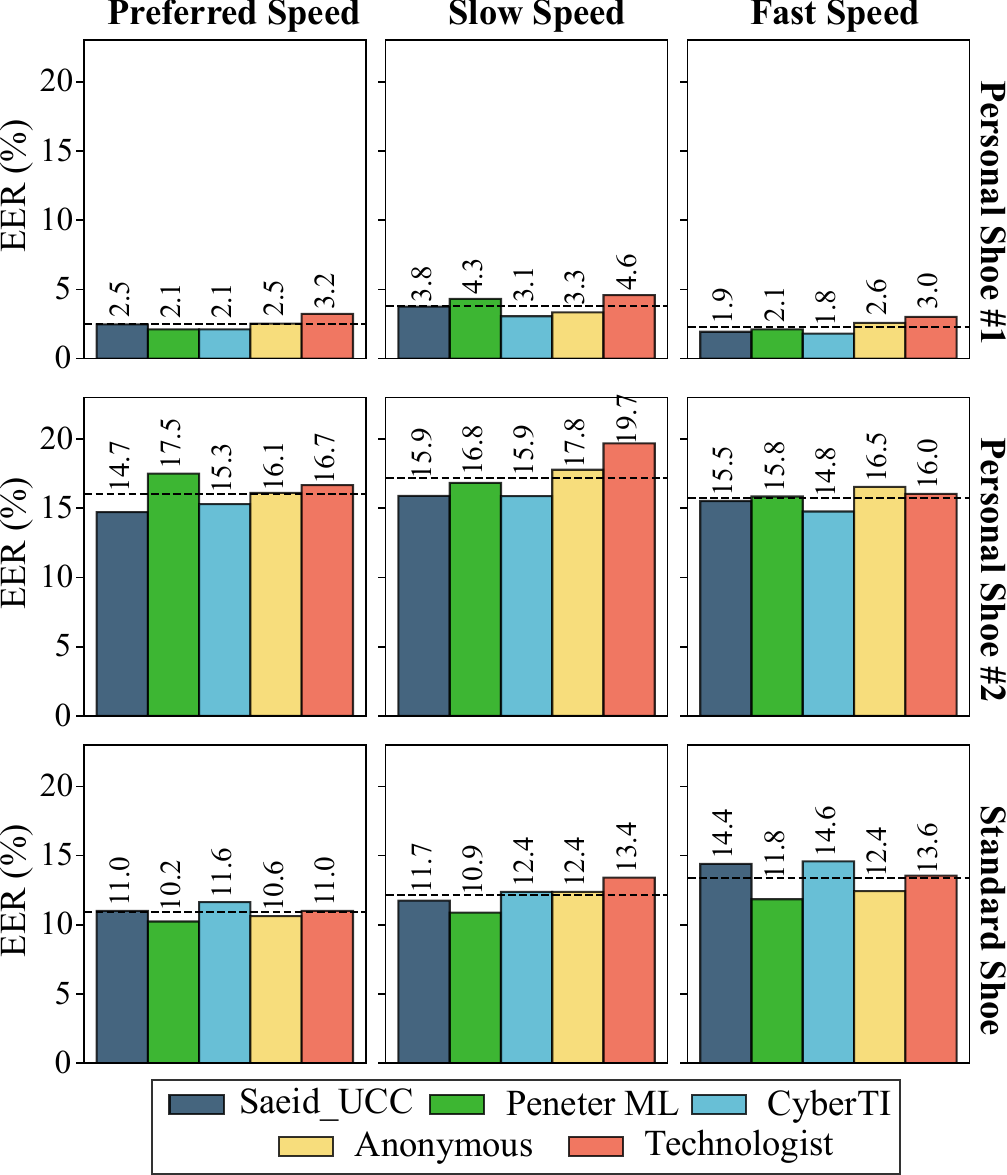}
    \caption{Performance (EER) for the top five submitted models across different walking speed and shoe type conditions. The Preferred Speed, Personal Shoe \#1 probe samples (top left corner) include conditions `seen' in the reference set, while the others are `unseen' conditions.  
    }
    \label{fig:bar_plot}
\end{figure}
%%%%%%%%%%%%%%%%%%%%%%%%%%%%%%%%%%%%%%%%%%%%%%%%%%%%%%%%%%%

% overall performance
Table \ref{tab:leaderboard} shows the performance metrics for the top five teams, which achieved EERs within 1.6\% of one another and a clear margin (more than 2\%) ahead of the rest of the competitors.
The winning solution, submitted by the Saeid\_UCC team, had an impressive EER of 10.77\%, outperforming the runner-up by more than 0.5\%. 
For comparison, the na\"ive starter kit implementation provided to competitors yielded an EER of 19.5\%.
The detection error tradeoff (DET) curve in Fig. \ref{fig:DET} further characterizes each model's performance across different decision thresholds, including those selected by the competitors.
The solutions exhibited different tradeoff characteristics; in particular, Saeid\_UCC’s winning model performed better for security-focused operating points (i.e., low FMR) but poorer for convenience-focused operating points (i.e., low FNMR) compared to the other solutions.
These tradeoffs are important to consider in practice, as different applications may prioritize lower FMR or FNMR depending on their security and usability requirements. 
Notably, the chosen thresholds for all of the top teams landed on the higher FMR and lower FNMR side of the EER.
%% add more? Could not see anything that jumped out from the score distributions. 

% performance across different conditions
Some trends in performance emerged across the submitted models.
Figure~\ref{fig:bar_plot} illustrates each model's EER on probe samples stratified by walking speed and shoe type.
% negative (false-claim) samples from held-out individuals also stratified by condition, however, these are technically unseen.   
As expected, the models were able to classify samples derived from familiar conditions (i.e., footwear and walking speeds similar to those represented in the reference set) with high accuracy, with EERs ranging between 2.1\% and 3.2\% across models.
Generalization to unfamiliar walking speeds, while keeping shoe type constant, was also strong. 
Performance dropped slightly for slow walking (EERs 3.1\% - 4.6\%), but actually improved in some cases for fast walking relative to preferred speed walking (EERs 1.9\% - 3.0\%).
In contrast, generalization to unfamiliar footwear proved significantly more challenging, corroborating previously reported results \cite{larracy_stepgan}. 
The largest performance drops were observed when evaluating the participants' second pair of personal footwear, with EERs ranging from 14.7\% to 19.7\%.
For samples collected in the standard shoes, which were flat-soled sneakers common to all thirty individuals in the competition set, EERs ranged from 10.2\% to 14.6\%.
As footsteps collected in these same shoes were also included for each of the 150 individuals in the training set, it is possible that the models were able to learn more robust features for this shoe type than for the other, previously unseen, personal shoes.
Although some of the personal shoes in the competition set may have coincided by chance with those worn by others in the training set, many of these shoe styles represented entirely new domains.

%%%%%%%%%%%%%%%%%%%%%%%%%%%%%%%%%%%%%%%%%%%%%%%%%%%%%%%
\begin{figure}[tb!]
    \centering
    \includegraphics[width=0.6\linewidth]{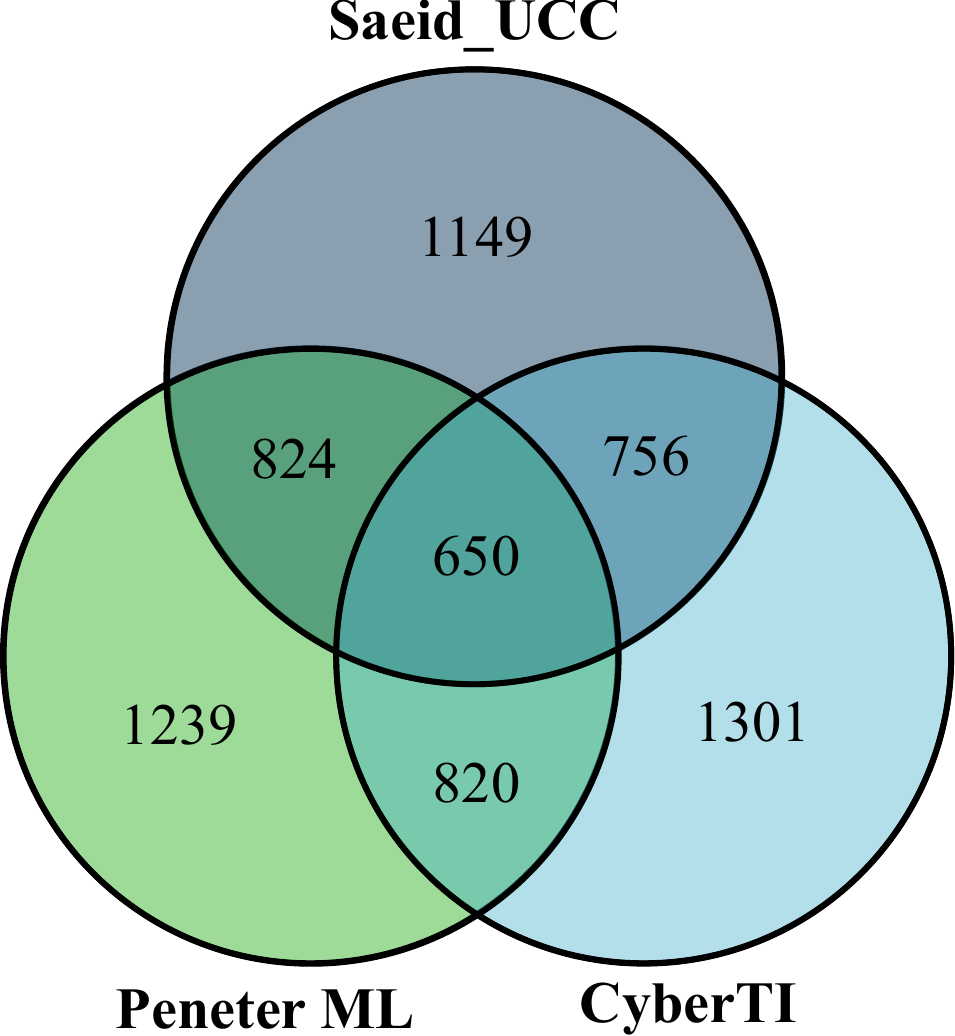}
    \caption{Overlap in misclassified probe samples for the top three solutions. There were 650 especially challenging footsteps that were misclassified by all three teams.}
    \label{fig:venn}
\end{figure}
%%%%%%%%%%%%%%%%%%%%%%%%%%%%%%%%%%%%%%%%%%%%%%%%%%%%%%%
\begin{figure}[tb!]
    \centering
    \includegraphics[width=\linewidth]{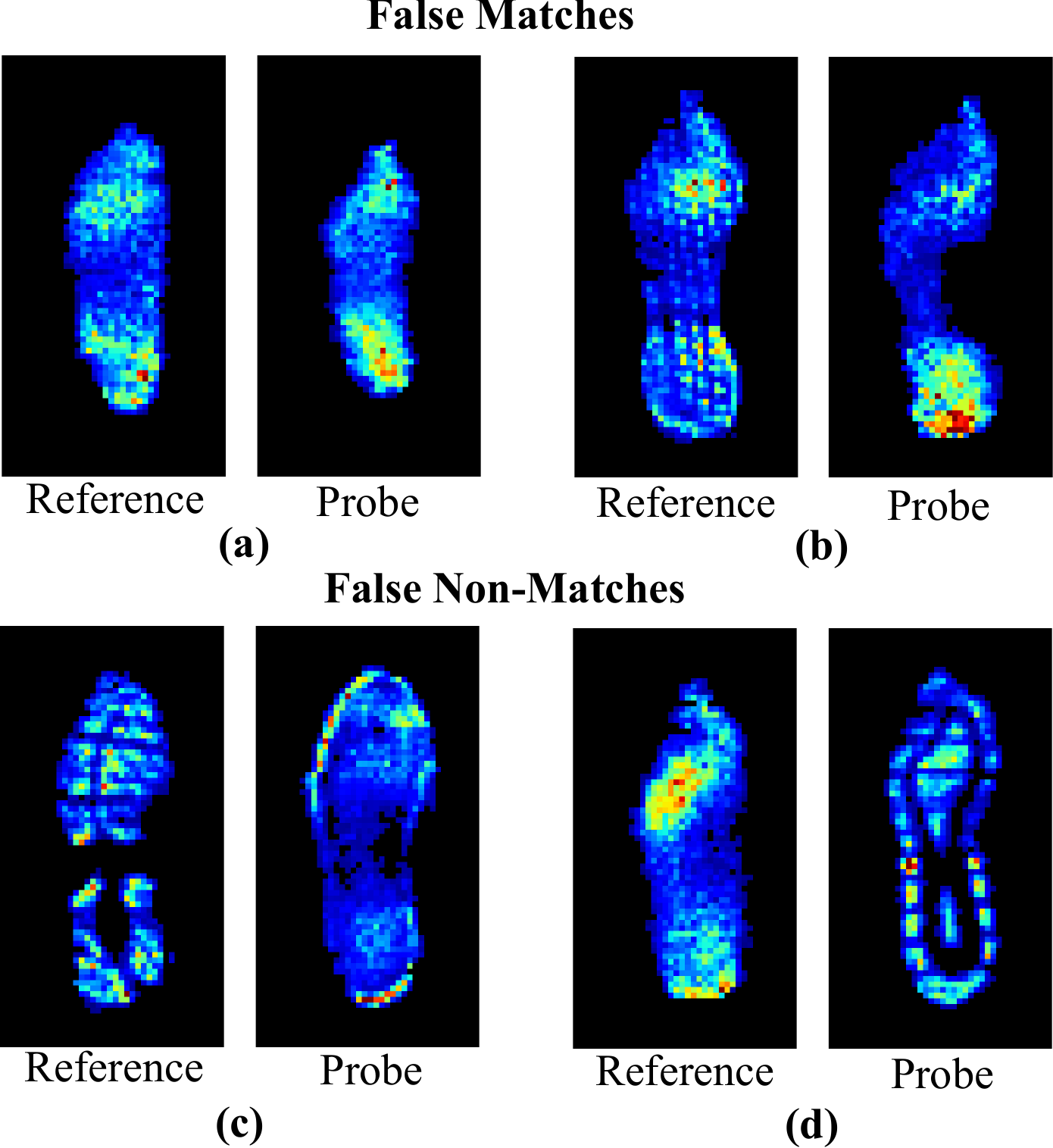}
    \caption{Examples of probe footsteps that were misclassified by all three of the top solutions, shown alongside reference samples for the claimed identities. Panels (a) and (b) depict false matches, and (c) and (d) depict false non-matches. Note: the 3D (time, height, width) footstep samples are shown here as 2D peak pressure images for visualization.}
    \label{fig:example_misclassed}
\end{figure}
%%%%%%%%%%%%%%%%%%%%%%%%%%%%%%%%%%%%%%%%%%%%%%%%%%%%%%%%%%%

% Misclassified samples
Figure~\ref{fig:venn} is a Venn diagram of the probe samples that were misclassified by each of the top three solutions, based on the decision thresholds provided by the competitors.
There were 650 probe samples that were misclassified by all three: among these, 175 (26.9\%) were false non-matches and 475 (73.1\%) were false matches, roughly reflecting the distribution of true and false identity claims in the probe set. 
Figure~\ref{fig:example_misclassed} shows examples of these particularly challenging cases, which were primarily related to variations in footwear. 
Notably, 53 false matches (of the total 475 false matches common to all three top solutions) originated from a single held-out participant who wore Birkenstock brand slip-on sandals, which were the same style as those worn by two enrolled users in their reference samples (see an example in Fig.~\ref{fig:example_misclassed}(a)). 
Similarly, 82 probe samples from one enrolled user were incorrectly rejected as non-matches by all three teams. 
These samples were collected in athletic shoes, representing a significant domain shift from the slip-on sandals worn in the user's reference recordings (see Fig.~\ref{fig:example_misclassed}(d)). 
These cases highlight persistent challenges in footstep biometrics caused by footwear, underscoring the need for further research.

%% summary of approaches
Despite the exploration of various architectures, including ViT, Swin, CNNLSTM, and ConvNeXt models, each of the top-performing teams ultimately selected the provided baseline R(2+1)D model as their backbone architecture. 
Originally developed for action recognition \cite{Tran2018}, the R(2+1)D CNNs proved to be well-suited for the aligned footstep data, and their relatively simple architecture may have supported stable training and efficient optimization.
Notably, the winning team opted for a lighter-weight variant of the network with fewer residual blocks and a more aggressively downsampled input than other solutions, highlighting that architectural simplicity, when combined with appropriate hyperparameters and a representative high-variability dataset, can be effective. 
Moreover, this solution incorporated data augmentation and a triplet loss function, which may have contributed to its superior performance; this points to promising directions for further exploration in combining augmentation techniques with deep metric learning to improve robustness to unseen conditions.

%% hyperparameter optimization strategies
Of note, all three top teams achieved performance gains through advanced optimization strategies that automated the selection of model architectures, hyperparameters, and/or training procedures. 
Rather than relying on hand-crafted features or domain-specific heuristics, these solutions used intelligent search frameworks to identify model configurations capable of generalizing across the high-variability samples.
Their success reflects not only the strength of the methods but also the value of the StepUP-P150 dataset: its diversity enabled effective optimization without overfitting, supporting robust performance on unseen data.
Moreover, as the specific impact of factors like footwear and walking speed on footstep recognition are not yet well understood, this data-driven approach may have offered a more reliable and immediate path to improvement than architectural innovation at this stage.
Nevertheless, these results highlight opportunities for further progress by pairing automated optimization with task-informed model design, including architectures and loss functions tailored to promote invariance to key covariates.

\section{Conclusion} 

The First International StepUP Competition for Biometric Footstep Recognition marked an important milestone for the field, celebrating the newly released UNB StepUP-P150 dataset and establishing a new and challenging benchmark. 
The competition attracted global participation and showcased promising strategies for improving footstep verification under real-world variability. 
While several submissions demonstrated strong performance, particularly under familiar conditions, all top-performing models struggled with generalization to unseen footwear, emphasizing the need for continued research. 
The competition’s CodaBench site will remain open for submissions, and researchers are encouraged to continue proposing new solutions to surpass the top teams presented here.
We hope this competition serves as a catalyst for further advancements in footstep biometrics and contributes meaningfully toward the development of practical, deployable systems.

\section*{Acknowledgments}
This project was supported by the New Brunswick Innovation Foundation's Strategic Opportunities Fund, the Atlantic Canada Opportunities Agency's Regional Innovation Ecosystem program, and the Natural Sciences and Engineering Research Council of Canada's Alliance Grants program [ALLRP 558340-20].
The competition organizers would like to thank all of the competitors that participated in the challenge. 

{\small
\bibliographystyle{ieee}
\bibliography{egbib}
}

\end{document}